\documentclass[twoside]{article}

\usepackage{natbib}

\usepackage{bm} 
\usepackage{amssymb}

\usepackage{booktabs}
\usepackage{url}

\usepackage{color}
\usepackage{amsmath,amsthm,amsfonts}
\usepackage{graphicx}

\renewenvironment{abstract}{\vskip.075in\centerline{\large\bf
Abstract}\vspace{0.5ex}\begin{quote}}{\par\end{quote}\vskip 1ex}

\title{Gaussian Process Regression Networks}

\author{\begin{tabular}{ccc}
    {Andrew Gordon Wilson\thanks{http:\///mlg.eng.cam.ac.uk\//andrew}} \qquad& David A. Knowles\thanks{http://mlg.eng.cam.ac.uk/dave} \qquad& Zoubin Ghahramani\thanks{http://mlg.eng.cam.ac.uk/zoubin}\\
     University of Cambridge  & University of Cambridge & University of Cambridge \\ 
   {\tt agw38@cam.ac.uk} & {\tt dak33@cam.ac.uk} & {\tt zoubin@eng.cam.ac.uk}
  \end{tabular}
}

\begin{document}
\date{}

\maketitle

\begin{abstract}
  We introduce a new regression framework, Gaussian process regression networks (GPRN),
  which combines the structural properties of Bayesian neural networks with the 
  nonparametric flexibility of Gaussian processes.  This model accommodates input 
  dependent signal and noise correlations between multiple response variables, input dependent 
  length-scales and amplitudes, and heavy-tailed predictive distributions.  We derive both efficient Markov 
  chain Monte Carlo and variational Bayes inference procedures for this model.  We apply GPRN as
  a multiple output regression and multivariate volatility model, demonstrating 
  substantially improved performance over eight popular multiple output (multi-task) Gaussian
  process models and three multivariate volatility models on benchmark datasets, including
  a 1000 dimensional gene expression dataset.
\end{abstract}

\section{Introduction}

Gaussian process models have become exceptionally popular for solving non-linear regression and
classification problems.  They are expressive, interpretable, avoid over-fitting, and have
impressive predictive performance in many thorough empirical comparisons 
\citep{rasmussenphd96, kuss2005, rasmussen06}.

In machine learning, Gaussian process regression developed out of neural networks research.  
\citet{neal1996} showed that Bayesian neural networks became Gaussian processes as the number of hidden 
units approached infinity, and conjectured that ``there may be simpler ways to do inference in this case.''  
These simple inference techniques became the cornerstone of subsequent Gaussian process models.  
However, neural networks had been motivated in part by their ability to capture correlations between 
multiple outputs (responses), by using adaptive hidden units that were shared between the outputs. 
In the infinite limit, this ability was lost.  

Recently there has been an explosion of interest in extending the Gaussian process regression framework
to account for \textit{fixed} correlations between output variables 
\citep{alvarez2011, yu2009, alvarez2008, bonilla2008, osborne2008, teh2005, boyle2004}.
These are often called `multi-task' learning  or `multiple output' regression models.
Capturing correlations between outputs (response variables) can be used to make 
better predictions.  Imagine we wish to predict cadmium concentrations in a region
of the Swiss Jura, where geologists are interested in heavy metal concentrations.  
A standard Gaussian process regression model would only be able to use cadmium training measurements.  
With a multi-task method, we can also make use of correlated 
heavy metal measurements to enhance cadmium predictions \citep{goovaerts1997}.  We
could further enhance predictions if we make use of how these (signal) correlations
change with geographical location.

There has similarly been great interest in extending Gaussian process (GP) regression 
to account for input dependent noise variances 
\citep{goldberg98,kersting2007,adams2008,turner2008,turner10,wilson2010,wilson2010gwp,lazaro2011}.
\citet{wilson2010gwp,wilson2011gwp} and \citet{fox2011bayesian} further extended the GP framework to accommodate input dependent 
noise correlations between multiple output (response) variables.

Other extensions include Gaussian process regression with non-stationary covariance function amplitudes 
\citep{turner2008,adams2008} and length-scales \citep{gibbs97,schmidt2003bayesian}, and with 
heavy tailed predictive distributions \citep{neal1997,vanhatalo09} for outlier 
rejection \citep{finetti1956,dawid1973,ohagan1979}.

In this paper, we introduce a new regression framework, Gaussian Process Regression Networks (GPRN),
which combines the structural properties of Bayesian neural networks \citep{neal1996} with the 
nonparametric flexibility of Gaussian processes.  This network is an adaptive mixture of Gaussian
processes, which naturally accommodates input dependent signal and noise correlations between 
multiple output variables, input dependent length-scales and amplitudes, and heavy tailed predictive 
distributions, without expensive or numerically unstable computations.  

We start by introducing the GPRN framework, and show how to perform efficient inference using both Markov 
chain Monte Carlo (MCMC) and variational Bayes (VB).  Carefully following \citet{alvarez2011}, we compare 
to eight multiple output GP models on gene expression and geostatistics datasets.  
We then compare to multivariate volatility models on several benchmark financial datasets, following
\citet{wilson2010gwp}.  In the Appendix, we review Gaussian process regression and the
notation of \citet{rasmussen06}.

\section{Gaussian Process Regression Networks}
\label{sec: gpn}

Given vector valued data points $\mathcal{D} = \{\bm{y}(x_i) : i = 1,\dots,N\}$,
where $x \in \mathcal{X}$ is an arbitrary input variable, we aim to predict 
$\mathbb{E}[\bm{y}(x_*)|x_*,\mathcal{D}]$ and $\text{cov}[\bm{y}(x_*)|x_*,\mathcal{D}]$
at a test input $x_*$.  We assume that the noise and signal correlations between the
elements of $\bm{y}(x)$ may change as a function of $x$.  Supposing $x$ is time 
($x=t$), a particular component of the $p$-dimensional vector $\bm{y}(t)$ could be the expression
level of a particular gene at time $t$, and $\Sigma(t)$ would then represent the variances 
and correlations for these genes at time $t$.  Instead of assuming only time dependence, we could 
have a more general input (predictor) variable $x \in \mathcal{X}$.  Then we can imagine 
$\bm{y}(x)$ representing different heavy metal concentrations at a geographical location 
$x \in \mathbb{R}^2$. These are two examples from the experiments in Section \ref{sec: experiments}.

We model $\bm{y}(x)$ as
\begin{equation}
 \bm{y}(x) = W(x)[\bm{f}(x) + \sigma_f \bm{\epsilon}] + \sigma_y \bm{z} \label{eqn: genmodel}
\end{equation}
where $\bm{\epsilon}$ and $\bm{z}$ are i.i.d. $\mathcal{N}(0,I)$ white noise\footnote{The distribution of $\bm{z}$ 
could be Student-$t$, Laplace, or something different. Using diagonal noise would also be a straightforward extension.},
$W(x)$ is a $p \times q$ matrix of independent Gaussian processes such that
$W(x)_{ij} \sim \mathcal{GP}(0,k_w)$, and $\bm{f}(x) = (f_1(x),\dots,f_q(x))^{\top}$ 
is a $q \times 1$ vector of independent GPs with $f_i(x) \sim \mathcal{GP}(0,k_{f_i})$.  
Each of the latent Gaussian processes in $\bm{f}(x)$ have additive Gaussian noise.  
Changing variables to include the noise $\sigma_f \bm{\epsilon}$ we let $\hat{f}_i(x) \sim \mathcal{GP}(0,k_{\hat{f}_i})$, 
where
\begin{equation}
k_{\hat{f}_i}(x,x') = k_{f_i}(x,x') + \sigma_f^2 \delta_{xx'} \,, \label{eqn: transformation}
\end{equation}
and $\delta_{xx'}$ is the Kronecker delta.  

\begin{figure}
\centering
\includegraphics[scale=1]{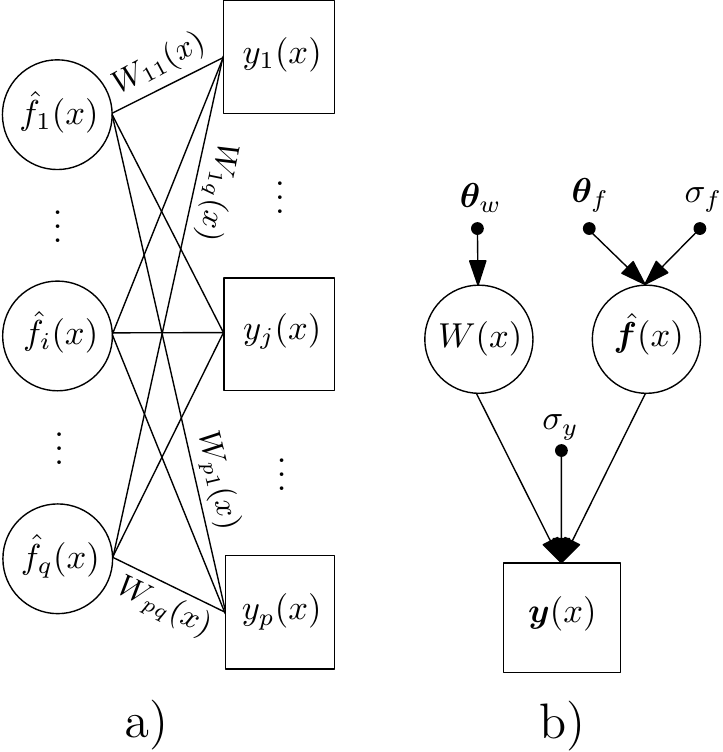}
\caption[Hallo]
{\small The Gaussian process regression network.  Latent random variables and 
observables are respectively labelled with circles and squares, except for the weight functions
in a).  Hyperparameters are labelled with dots.  a) This neural network style diagram shows 
the $q$ components of the vector $\hat{\bm{f}}$ (GPs with additive noise), and the $p$ components of the vector $\bm{y}$.  
The links in the graph, four of which are labelled, are latent random weight \textit{functions}.  Every 
quantity in this graph depends on the input $x$.  This graph emphasises the adaptive nature of this network: links can change strength or even disappear
as $x$ changes. b) A directed graphical model showing the generative procedure with relevant variables.}
\label{fig: gpnet}
\end{figure}

We represent this \textit{Gaussian process regression network} (GPRN) in Figure \ref{fig: gpnet}, labelling
the length-scale hyperparameters for the kernels $k_w$ and $k_f$ as $\bm{\theta}_w$ and
$\bm{\theta}_f$ respectively.  We see the latent \textit{node functions} $\hat{\bm{f}}(x)$ 
are connected together to form the outputs $\bm{y}(x)$.  The strengths of the connections 
change as a function of $x$; the weights themselves -- the entries of $W(x)$ -- are functions.  Old 
connections can break and new connections can form. This is an \textit{adaptive} network, where the signal 
and noise correlations between the components of $\bm{y}(x)$ vary with $x$.
\footnote{Coincidentally, there is an unrelated paper called ``Gaussian process networks'' \citep{friedman2000gaussian},
which is about learning the structure of Bayesian networks -- e.g. the direction of dependence between 
random variables.}  

To explicitly separate the dynamic signal and noise correlations, we re-write \eqref{eqn: genmodel} as
\begin{align}
 \bm{y}(x) = \underbrace{W(x)\bm{f}(x)}_{\text{signal}} + \underbrace{\sigma_f W(x)\bm{\epsilon} + \sigma_y \bm{z}}_{\text{noise}}  \,. \label{eqn: genmodelnew}
\end{align}

Conditioning on $W(x)$ in \eqref{eqn: genmodelnew}, we can better understand the signal correlations.  In this case, 
each of the outputs $\bm{y}_i(x)$, $i=1,\dots,p$, is a Gaussian process with kernel
\begin{equation}
k_{y_i}(x,x') = \sum_{j=1}^{q} W_{ij}(x) [k_{f_j}(x,x')+\sigma_f^2] W_{ij}(x') + \sigma_y^2 \,. \label{eqn: mixing}
\end{equation}
Even if $\sigma_f^2$ and $\sigma_y^2$ are zero, so that this is a noise free regression, 
there are still signal correlations; the components of $\bm{y}$ are coupled through 
the matrix $W(x)$.  Once the network has been trained, $W(x)$ is conditioned on the
data $\mathcal{D}$, and so the predictive covariances of $\bm{y}(x_*) | \mathcal{D}$ 
are now influenced by the values of the observations themselves, and not just distances 
between the test point $x_*$ and the observed points $x_1,\dots,x_N$ as is the case for 
independent GPs; we can view 
\eqref{eqn: mixing} as an adaptive kernel learned from the data.  There are three other 
interesting features in equation \eqref{eqn: mixing}:  
1) the amplitude of the covariance function $\sum_{j=1}^{q} W_{ij}(x)W_{ij}(x')$ 
is non-stationary (input dependent); 2) even if each of the kernels $k_{f_j}$ has different
\textit{stationary} length-scales, the mixture of the kernels $k_{f_j}$ is input dependent and
so the effective overall length-scale is non-stationary;
3) the kernels $k_{f_j}$ may be entirely different: some may be periodic, others
squared exponential, others Brownian motion, etc. . So the overall covariance 
function (kernel) may be continuously switching between regions of entirely
different covariance structures.

In addition to modelling signal correlations, we can see from equation \eqref{eqn: genmodelnew} 
that the GPRN is simultaneously a multivariate volatility
model.  The noise covariance is $\sigma_f^2 W(x) W(x)^{\top} + \sigma_y^2 I$.  Since the entries of $W(x)$ are GPs, this noise model is an example of a \textit{generalised Wishart process} \citep{wilson2010gwp,wilson2011gwp}.  

The number of nodes $q$ influences how the model accounts for signal and
noise correlations.  If $q$ is smaller than $p$, the dimension 
of $\bm{y}(x)$, the model performs dimensionality reduction and matrix factorization
as part of the regression on $\bm{y}(x)$ and $\text{cov}[\bm{y}(x)]$.  However, we may  
want $q > p$, for instance if the output space were one dimensional ($p=1$).  In this case we would need $q > 1$ to realise features 2 and 3 listed above.  For a given dataset,
we can vary $q$ and select the value which gives the highest marginal likelihood on
training data.  We can also use `automatic relevance determination' \citep{mackay1994automatic} as a proxy for model selection for $q$ for a given dataset. 
This is achieved by introducing $\{a_j\}$, signal variances for each node function $j$, so that $k_{\hat{f}_j} \to a_j k_{\hat{f}_j}$,
and comparing magnitudes of the trained $a_j$.

When $q = p = 1$, the GPRN essentially becomes the nonstationary 
GP regression model of \citet{adams2008} and \citet{turner2008}.
Likewise, when the weight functions are constants the 
GPRN becomes the Semiparametric Latent Factor Model    
(SLFM) of \citet{teh2005}, except that the resulting GP regression network is less 
prone to over-fitting through its use of full Bayesian inference.\footnote{In \citet{teh2005},
the weight constants are a large matrix of hyperparameters, determined through 
maximising a marginal likelihood.}  Indeed, in an implementation of GPRN, 
one can switch features on or off; to switch off changing correlations and multivariate 
volatility, set $\sigma_f^2 = 0$ and the length-scales for weight function kernels ($k_w$) to large 
fixed values.

\section{Inference}
\label{sec: inference}

Now that we have specified a prior $p(\bm{y}(x))$ at all points $x$ in our domain $\mathcal{X}$, we wish to
predict $\mathbb{E}[\bm{y}(x_*)|x_*,\mathcal{D}]$ and $\text{cov}[\bm{y}(x_*)|x_*,\mathcal{D}]$ 
at a test input $x_*$, given vector valued data $\mathcal{D} = \{\bm{y}(x_i) : i = 1,\dots,N\}$. We do this using two different approaches -- variational Bayes and Markov chain Monte Carlo (MCMC) -- and we compare between 
these approaches.  We also use variational Bayes to estimate all hyperparameters 
$\bm{\gamma} = \{\bm{\theta}_f,\bm{\theta}_w,\sigma_f,\sigma_y\}$, where, as before, 
$\bm{\theta}_f$ and $\bm{\theta}_w$ are the length-scales of the node and weight function kernels.

As a first step, we re-write the prior in terms of $\bm{u}=(\hat{\mathbf{f}}, \mathbf{W})$, a vector composed of 
all the node and weight Gaussian process functions, evaluated at the training points $\{x_1,\dots,x_N\}$.  
There will be $q$ node functions and $p \times q$ weight functions.  Therefore
\begin{equation}
 p(\bm{u}|\sigma_f,\bm{\theta}_f,\bm{\theta}_w) = \mathcal{N}(0,C_B) \,, 
\end{equation}
where $C_B$ is an $Nq(p+1)\times Nq(p+1)$ block diagonal matrix, since the weight and 
node functions are independent in the prior.  The way we have ordered $\bm{u}$,
the first $q$ blocks are $N \times N$ covariance matrices $K_{\hat{f}}$ from the node kernel 
$k_{\hat{f}}$, and the last blocks are $N \times N$ covariance matrices $K_w$ from the 
weight kernel $k_{w}$.

Next we specify our likelihood function, so we can use Bayes' theorem to find the posterior 
$p(\bm{u}|\mathcal{D},\bm{\gamma})$. From \eqref{eqn: genmodel}, our likelihood is
\begin{equation}
 p(\mathcal{D} | \bm{u},\sigma_y) = \prod_{i=1}^{N} \mathcal{N}(\bm{y}(x_i); W(x_i)\hat{\bm{f}}(x_i),\sigma_y^2 I) \,.  \label{eqn: likelihood}
\end{equation}
By incorporating noise on $\bm{f}$, the GP network accounts for multivariate volatility (as in \eqref{eqn: genmodelnew}),
without the need for costly or numerically unstable 
matrix inversions.  For other multivariate volatility models,
like multivariate GARCH \citep{bollerslev88}, or multivariate stochastic volatility \citep{harvey94}, 
the likelihood takes the form $p(\mathcal{D}| \beta) =  \prod_{i=1}^{N} \mathcal{N}(\bm{\mu}_i,\Sigma_i)$, 
and requires inversions of $p \times p$ covariance matrices. 
There are three other notable advantages to the inference
with GPRN: 1) it is easy to simultaneously estimate $\bm{\mu}_i$ and $\Sigma_i$.  Usually in the multivariate
volatility setting, $\bm{\mu}_i$ is assumed to be a constant; 2) we can use a Student-$t$ observation model
instead of a Gaussian observation model, by letting $\bm{z}$ in \eqref{eqn: genmodel} be $t$ distributed, 
with minimal changes to the inference procedures; 3) we can transform the components of the product $W(x_i)\hat{\bm{f}}(x_i)$ so that the
priors on the components of $\bm{y}(x_i)$ become \textit{copula processes} \citep{wilson2010} and have whatever
marginals we desire.  We can also do this without significantly changing inference procedures.  

Now that we have specified our prior and likelihood, we can apply Bayes' theorem:
\begin{align}
 p(\bm{u}|\mathcal{D},\bm{\gamma}) \propto p(\mathcal{D}|\bm{u},\sigma_y)p(\bm{u}|\bm{\theta}_f,\bm{\theta_w},\sigma_f) \,. \label{eqn: posterior}
\end{align}
In the next sections, we discuss how to either sample from or use variational Bayes to approximate
this posterior in \eqref{eqn: posterior}, so that we can estimate $p(\bm{y}(x_*)|\mathcal{D})$.
We also use variational Bayes to learn the hyperparameters $\bm{\gamma}$.

\subsection{Markov chain Monte Carlo}
\label{sec: mcmc}

To sample from \eqref{eqn: posterior}, we could use a Gibbs 
sampling scheme which would have conjugate posterior updates,
alternately conditioning on weight and node functions.  
However, this Gibbs cycle would mix poorly because of the
tight correlations between the weights and the nodes.  In general, 
MCMC samples from \eqref{eqn: posterior} mix poorly because of the strong 
correlations in the prior imposed by $C_B$.  The sampling process is also often 
slowed by costly matrix inversions in the likelihood.  

We use Elliptical Slice 
Sampling \citep{murray10}, a recent MCMC technique specifically 
designed to sample from posteriors with tightly correlated Gaussian priors.  
It does joint updates and has no free parameters.  We find that it mixes
well.  And since there are no costly or numerically unstable matrix 
inversions in the likelihood of \eqref{eqn: likelihood} we also 
find sampling to be highly efficient.  

With a sample from \eqref{eqn: posterior}, we can sample from 
the predictive  
$p(W(x_*),{\bm{f}}(x_*)|\bm{u},\sigma_f,\mathcal{D})$.  Let
$W_*^i,{\bm{f}}_*^i$ be the $i^{\text{th}}$ such 
joint sample.  Using \eqref{eqn: genmodelnew} we can then 
construct samples of $p(\bm{y}(x_*)|W_*^i,\bm{f}_*^i,\sigma_f,\sigma_y)$,
from which we can construct the predictive distribution
\begin{equation}
p(\bm{y}(x_*)|\mathcal{D}) = \lim_{J \to \infty} \frac{1}{J} \sum_{i=1}^{J} p(\bm{y}(x_*)|W_*^i,\bm{f}_*^i,\sigma_f,\sigma_y) \,. \label{eqn: predictive}
\end{equation}
We see that even with a Gaussian observation model, the predictive distribution in \eqref{eqn: predictive} is an infinite mixture of 
Gaussians, and will generally be heavy tailed and therefore robust to outliers.  

Mixing was assessed by looking  at trace plots of samples, and the likelihoods of these samples.  
Specific information about how long it takes to sample a solution for a given problem is in the 
experiments section.  

\subsection{Variational Bayes}  
\label{sec: vb}

We perform variational EM \citep{jordan1999introduction} to fit an approximate posterior $q$ to the true posterior $p$, by minimising the Kullback-Leibler divergence $KL(q||p)=
  -H[q(\mathbf{v})] - \int q(\mathbf{v}) \log p(\mathbf{v}) d\mathbf{v},
$
where $H[q(\mathbf{v})]=-\int q(\mathbf{v}) \log q(\mathbf{v}) d\mathbf{v}$ is the entropy and $\mathbf{v}=\{\mathbf{f}, \mathbf{W}, \sigma^2_{f},\sigma^2_y,a_j\}$. 

\paragraph{E-step.} We use Variational Message Passing~\citep{Winn2006} under the Infer.NET framework~\citep{InferNET10} to estimate the posterior over $\mathbf{v}=\{\mathbf{f}, \mathbf{W}, \sigma^2_{f},\sigma^2_y,a_j\}$.  We specify inverse Gamma priors on $\{\sigma^2_{f},\sigma^2_y,a_j\}$:
\begin{align*}
\sigma^2_{fj} \sim \text{IG}(\alpha_{\sigma^2_{f}},\beta_{\sigma^2_{f}}), \ 
\sigma^2_y \sim \text{IG}(\alpha_{\sigma^2_y},\beta_{\sigma^2_y}), \ 
a_j\sim \text{IG}(\alpha_{a},\beta_{a}). 
\end{align*}
For mathematical and computational convenience we introduce the following variables which are deterministic functions of the existing variables in the model:
\begin{align}
w_{nij}&:=W_{ij}(x_n), \qquad
f'_{nj}:=f_j(x_n) \\
t_{nij}&:=w_{nij}\hat{f}_{nj}, \qquad
s_{in}:=\sum_j t_{nij} 
\end{align}
Note that the observations $y_i(x_n) \sim \mathcal{N}(s_{in}, \sigma_y^2)$ and that $\hat{f}_{nj}\sim \mathcal{N}(f'_{nj},\sigma^2_{f_j})$. Variational message passing uses these deterministic factors and the associated ``pseudo-marginals'' as conduits to pass appropriate moments, resulting in the same updates as standard VB~\citep{Winn2006}. The full model can now be written as
\begin{align*}
p(\mathbf{v}) \propto \text{IG}(\sigma^2_y; \alpha_{\sigma^2_y},\beta_{\sigma^2_y})
\prod_{j=1}^Q \Biggl( \mathcal{N}(\mathbf{f}_j;0,a_jK_{f_j})  \\
\text{IG}(\sigma^2_{fj};\alpha_{\sigma^2_{f}},\beta_{\sigma^2_{f}})
\text{IG}(a_j; \alpha_{a},\beta_{a})
  \prod_{i=1}^P \Biggl[ \mathcal{N}(\mathbf{W}_{ij};0,K_w)  \\  
  \prod_{n=1}^N \delta(w_{nij}-W_{ij}(x_n))
\delta(f'_{nj}-\hat{f}_j(x_n))
\mathcal{N}(\hat{f}_{nj}; f'_{nj},\sigma^2_{f_j})   \\
\delta(t_{nij}-w_{nij}\hat{f}_{nj})
\delta(s_{in}-\sum_j t_{nij})
\mathcal{N}(y_i(x_n);s_{in}, \sigma_y^2) \Biggr] \Biggr)
\end{align*}
We use a variational posterior of the following form:
\begin{align*}
q(\mathbf{v}) = q_{\sigma^2_y}(\sigma^2_y)
\prod_{j=1}^Q q_{\mathbf{f}_j}(\mathbf{f}_j) 
q_{\sigma^2_{fj}}(\sigma^2_{fj})
q_{a_j}(a_j)
  \prod_{i=1}^P q_{\mathbf{W}_{ij}}(\mathbf{W}_{ij}) \\ \prod_{n=1}^N q_{w_{nij}}(w_{nij})
q_{f'_{nj}}(f'_{nj})
q_{\hat{f}_{nj}}(\hat{f}_{nj})  
q_{t_{nij}}(t_{nij})
q_{s_{in}}(s_{in})
\end{align*}
where $q_{\sigma^2_y}, q_{\sigma^2_{fj}}$ and $q_{a_j}$ are inverse Gamma distributions; $q_{w_{nij}}, q_{f'_{nj}}, q_{\hat{f}_{nj}}, q_{t_{nij}}$ and $q_{s_{in}}$ are univariate normal distributions; and  $q_{\mathbf{f}_j}$ and $q_{\mathbf{W}_{ij}}(\mathbf{W}_{ij})$ are multivariate normal distributions. 

The updates for $\mathbf{f}, \mathbf{W}, \sigma^2_{f},\sigma^2_y$ are standard VB updates and are available in Infer.NET. The update for the ARD parameters $a_j$ however required specific implementation. The factor itself is 
\begin{align}
\log \mathcal{N}(\mathbf{f}_j;& \mathbf{0}, a_j K_f) \stackrel{c}{=} -\frac12 \log |a_jK_j|  - \frac12 \mathbf{f}_j^T (a_jK_f)^{-1} \mathbf{f}_j \nonumber \\ 
&= - \frac{N}2 \log a_j - \frac12 \log |K_j|  - \frac12 a_j^{-1} \mathbf{f}_j^T K_f^{-1} \mathbf{f}_j 
\label{eqn:gpfactor}
\end{align}
where $\stackrel{c}{=}$ denotes equality up to an additive constant. Taking expectations with respect to $\mathbf{f}$ under $q$ we obtain the VMP message to $a_j$ as being $\text{IG}\left(a_j;\frac{N}2  - 1, \frac12 \langle \mathbf{f}_j^T K_f^{-1} \mathbf{f}_j \rangle \right)$. Since the variational posterior on $\mathbf{f}$ is multivariate normal the expectation $\langle \mathbf{f}_j^T K_f^{-1} \mathbf{f}_j \rangle$ is straightforward to calculate. 

\paragraph{M-step.} In the M-step we optimise the variational lower bound with respect to the log length scale parameters $\{\theta_f,\theta_w\}$, using gradient descent with line search. When optimising $\theta_f$ we only need to consider the contribution to the lower bound of the factor $\mathcal{N}(\mathbf{f}_j;0,a_jK_{f_j})$ (see $\eqref{eqn:gpfactor}$), which is straightforward to evaluate and differentiate (see Appendix). For $\theta_w$ we consider the contribution of $\mathcal{N}(\mathbf{W}_{pq};0,K_{W})$. 

\subsection{Computational Considerations}
\label{sec: computation}
GPRN is mainly limited by taking the Cholesky 
decomposition of the block diagonal $C_B$, an $Nq(p+1)\times Nq(p+1)$ matrix.  But  
$pq$ of these blocks are the same $N \times N$ covariance matrix $K_w$ for the weight
functions, and $q$ of these blocks are the covariance matrices $K_{\hat{f}_i}$ 
associated with the node functions, and 
$\text{chol}(\text{blkdiag}(A,B,\dots) )= \text{blkdiag}(\text{chol}(A),\text{chol}(B),\dots)$.
Therefore assuming the node functions share the same covariance function 
(which they do in our experiments), the complexity of this operation is only $\mathcal{O}(N^3)$,
the same as for regular Gaussian process regression.  At worst it is $\mathcal{O}(qN^3)$, 
assuming different covariance functions for each node.  

Sampling also requires likelihood evaluations.  Since there are input dependent noise correlations
between the elements of the $p$ dimensional observations $\bm{y}(x_i)$, multivariate volatility
models would normally require inverting a $p \times p$ covariance matrix $N$ times, like  
MGARCH \citep{bollerslev88} or multivariate stochastic volatility models \citep{harvey94}.
This would lead to a total complexity of $\mathcal{O}(Nqp + Np^3)$.  However, by working 
directly with the noisy $\hat{\bm{f}}$ instead of the noise free $\bm{f}$, evaluating the 
likelihood requires no costly or numerically unstable inversions, and thus has a complexity of only 
$\mathcal{O}(Nqp)$.  This allows GPRN to scale to high dimensions; 
indeed we have a 1000 dimensional gene expression experiment in Section 
\ref{sec: experiments}.  

The computational complexity of VB is dominated by the $\mathcal{O}(N^3)$ inversions required to 
calculate the covariance of the node and weight functions in the E-step. Naively $q$ and $qp$ such 
inversions are required per iteration for the node and weight functions respectively, giving a total 
complexity of $\mathcal{O}(qpN^3)$. However, under VB the covariances of the weight functions for the 
same $p$ are all equal, reducing the complexity to $\mathcal{O}(qN^3)$. If $p$ is large the $\mathcal{O}(pqN^2)$ 
cost of calculating the weight function means may become significant. Although the per iteration cost of VB is actually 
higher than for MCMC far fewer iterations are typically required to reach convergence. 

Overall, even though the GPRN accounts for input dependent signal correlations (rather than
fixing the correlations like other multi-task methods), the computational demands of GPRN compare favourably to most 
multi-task GP models, which commonly have a complexity of $\mathcal{O}(p^3 N^3)$ \citep{alvarez2011}.

\section{Experiments}
\label{sec: experiments}

We compare the GPRN to multi-task learning and multivariate volatility 
models, and we also use the GPRN to gain new scientific insights into the data we model.
Furthermore, we compare between variational Bayes (VB) and Markov chain Monte Carlo (MCMC) inference
within the GPRN framework.  To keep our comparisons up to date, we \textit{exactly} reproduce many of the experiments 
in recent papers by \citet{alvarez2011} and \citet{wilson2010gwp} on benchmark datasets. 
In the multi-task setting, there are $p$ dimensional observations $\bm{y}(x)$, and the goal is to 
use the correlations between the elements of $\bm{y}(x)$ to make better predictions of $\bm{y}(x_*)$,
for a test input $x_*$, than if we were to treat the dimensions independently.  A major difference between 
GPRN and alternative multi-task models is that the GPRN accounts for signal correlations that \textit{change} with $x$, rather than fixed 
correlations.  It also accounts for changing noise correlations (multivariate volatility). 

We compare to the following multi-task GP methods: 1) the linear model of coregionalisation (LMC) \citep{journel1978,goovaerts1997},
2) the intrinsic coregionalisation model (ICM) \citep{goovaerts1997}, 3) ordinary co-kriging \citep{cressie1993statistics,goovaerts1997,wackernagel2003multivariate},
4) the semiparametric latent factor model (SLFM) \citep{teh2005}, 5) convolved multiple output Gaussian processes (CMOGP) \citep{barry1996blackbox,ver1998constructing,boyle2004},
6) standard independent Gaussian processes (GP), 7) and the DTC \citep{csato2001sparse,seeger2003fast,quinonero2005unifying,rasmussen06} , 8) FITC \citep{snelson2006sparse}, 
and 9) PITC \citep{quinonero2005unifying} sparse approximations for CMOGP \citep{alvarez2011}, which we respectively label as
MDTC, MFITC and MPITC.  Detail about each of these methods
is in \citet{alvarez2011}.  We compare on 
a 3 dimensional geostatistics heavy metal dataset from the Swiss Jura, where 28\% of the 
observations for one of the outputs (response variables) is missing, and on gene expression
datasets with 50 and 1000 dimensional time dependent outputs $\bm{y}(t)$.

In the multi-task experiments, the GPRN accounts for input dependent noise covariance matrices
$\text{cov}[\bm{y}(x)] = \Sigma(x)$.  To specifically test GPRN's ability to model input dependent
noise covariances (multivariate volatility), we also compare predictions of $\Sigma(x)$ to those
made by popular multivariate volatility models -- full BEKK MGARCH \citep{engle1995}, generalised Wishart 
processes \citep{wilson2010gwp}, the original Wishart process \citep{bru91,gourieroux09}, and empirical estimates -- on  
benchmark return series datasets which are especially suited to MGARCH 
\citep{granger05, hansen05, engle2009, mccullough98, brooks2001}.

In all experiments, GPRN uses a squared exponential covariance function for its node functions, and another 
squared exponential covariance function for its weight functions.

\subsection{Gene Expression}

\citet{tomancak2002} measured gene expression levels every hour for 12 hours during Drosophila embryogenesis; they then
repeated this experiment for an independent replica (a second independent time series).  Gene expression is activated 
and deactivated by transcription factor proteins.  We focus on genes which are thought to at least be regulated by 
the transcription factor \texttt{twi}, which influences mesoderm and muscle development in Drosophila \citep{zinzen2009combinatorial}.  
The assumption is that these gene expression levels are all correlated.  We would like to use how these correlations 
change over time to make better predictions of time varying gene expression in the presence of transcription factors.
In total there are 1621 genes (outputs) at $N = 12$ time points (inputs), on two independent replicas.  For training,
$p=50$ random genes were selected from the first replica, and the corresponding 50 genes in the second replica were 
used for testing.  We then repeated this experiment 10 times with a different set of genes each time, and averaged
the results.  We then repeated the whole experiment, but with $p=1000$ genes.  We used \textit{exactly} the same training 
and testing sets as \citet{alvarez2011}.  

We use a relatively small $p=50$ dataset so that we are able to compare with popular alternative multi-task 
methods (LMC, CMOGP, SLFM) which have a complexity of $\mathcal{O}(N^3 p^3)$ and would not
scale to $p=1000$ \citep{alvarez2011}.  For $p=1000$, we compare to the sparse convolved multiple output GP methods (MFITC, MDTC, and MPITC) of \citet{alvarez2011}.  In both of these regressions, the GPRN is accounting 
for multivariate volatility; this is the first time a multivariate stochastic volatility model has been estimated for $p > 50$ \citep{chib2006}.  
We assess performance using standardised mean square error (SMSE) and mean standardized log loss (MSLL), as defined in \citet{rasmussen06} on page
23, and discussed in \citet{alvarez2011} on page 1469.  Using the empirical mean and variance to fit the data would
give an SMSE and MSLL of $1$ and $0$ respectively.  The smaller the SMSE and more negative the MSLL the better.  

The results are in Table \ref{tab: predictions}, under the headings \texttt{GENE} (50D) and \texttt{GENE} (1000D). 
For \texttt{SET 1} of the 50D dataset, we used Replica 1 and Replica 2 in \citet{alvarez2011} respectively as 
training and testing replicas.  We follow \citet{alvarez2011} and reverse training and testing 
replicas to create \texttt{SET 2}.  The results for LMC, CMOGP, MFITC, MPITC, and MDTC are reproduced from \citet{alvarez2011}.  
GPRN significantly outperforms all of the other models, with between
46\% and 68\% of the SMSE, and similarly strong results on the MSLL error metric.  On the 50D dataset the
MCMC and VB results are comparable.  However, on the 1000D dataset GPRN with VB noticeably outperforms 
GPRN with MCMC, likely because MCMC is not mixing as well in high dimensions.  Indeed VB may have an 
advantage over MCMC in high dimensions.  On the other hand, GPRN with MCMC is still robust, 
outperforming all the other methods on the 1000 dimensional dataset.

On both the 50 and 1000 dimensional datasets, the marginal likelihood for the network structure
is sharply peaked at $q=1$.  This is evidence for the hypothesis that there is only the one transcription 
factor \texttt{twi} controlling the expression levels of the genes in question.  These datasets can be found in Neil Lawrence's GPSIM toolbox: \url{http://staffwww.dcs.shef.ac.uk/people/N.Lawrence/gpsim/}

Typical GPRN (VB) runtimes for the 50D and 1000D datasets were respectively 12 seconds and 330 seconds.

\subsection{Jura Geostatistics} 

Here we are interested in predicting concentrations of cadmium at 100 locations within a  
14.5 km$^2$ region of the Swiss Jura.  For training, we have access to measurements of 
cadmium at 259 neighbouring locations.  We also have access to nickel and zinc 
concentrations at these 259 locations, as well as at the 100 locations we wish to 
predict cadmium.  While a standard Gaussian process regression model would only be able to make use of 
the cadmium training measurements, a multi-task method can use the correlated nickel and zinc measurements 
to enhance predictions\footnote{This can be seen as a multivariate missing data problem, with
$p=3$ outputs.}. With GPRN we can also make use of how the correlations between nickel, 
zinc, and cadmium change with location to further enhance predictions.  

\begin{figure}
\centering
\includegraphics{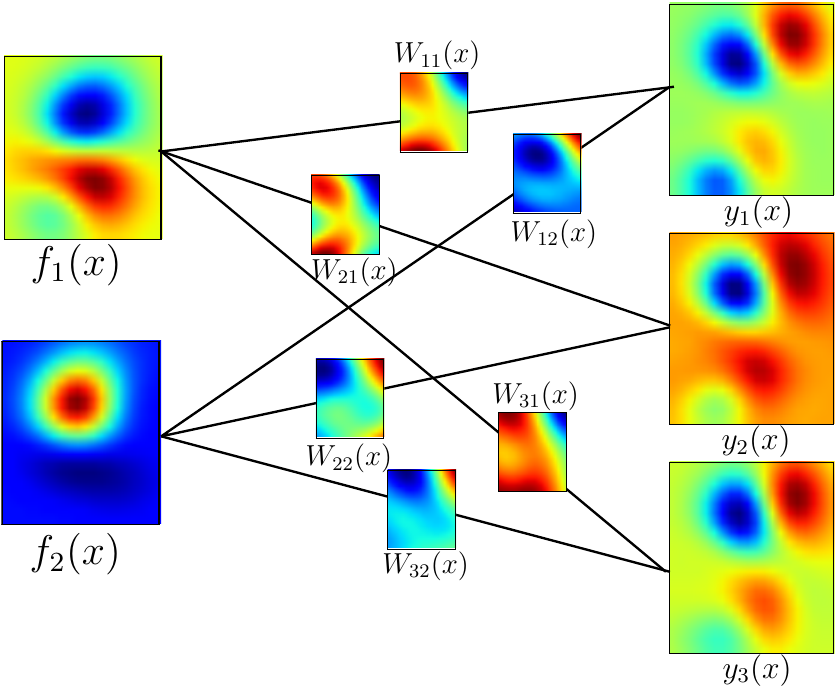}
\caption[Hallo]
{\small Network structure for the Jura dataset learnt by GPRN. The layout here is as in Figure~\ref{fig: gpnet}, with the spatially varying node and weight functions shown, along with the predictive means for the observations. The three output dimensions are cadmium, nickel and zinc concentrations respectively.}
\label{fig:juranetwork}
\end{figure}

\begin{figure}
\centering
\includegraphics[scale=0.55]{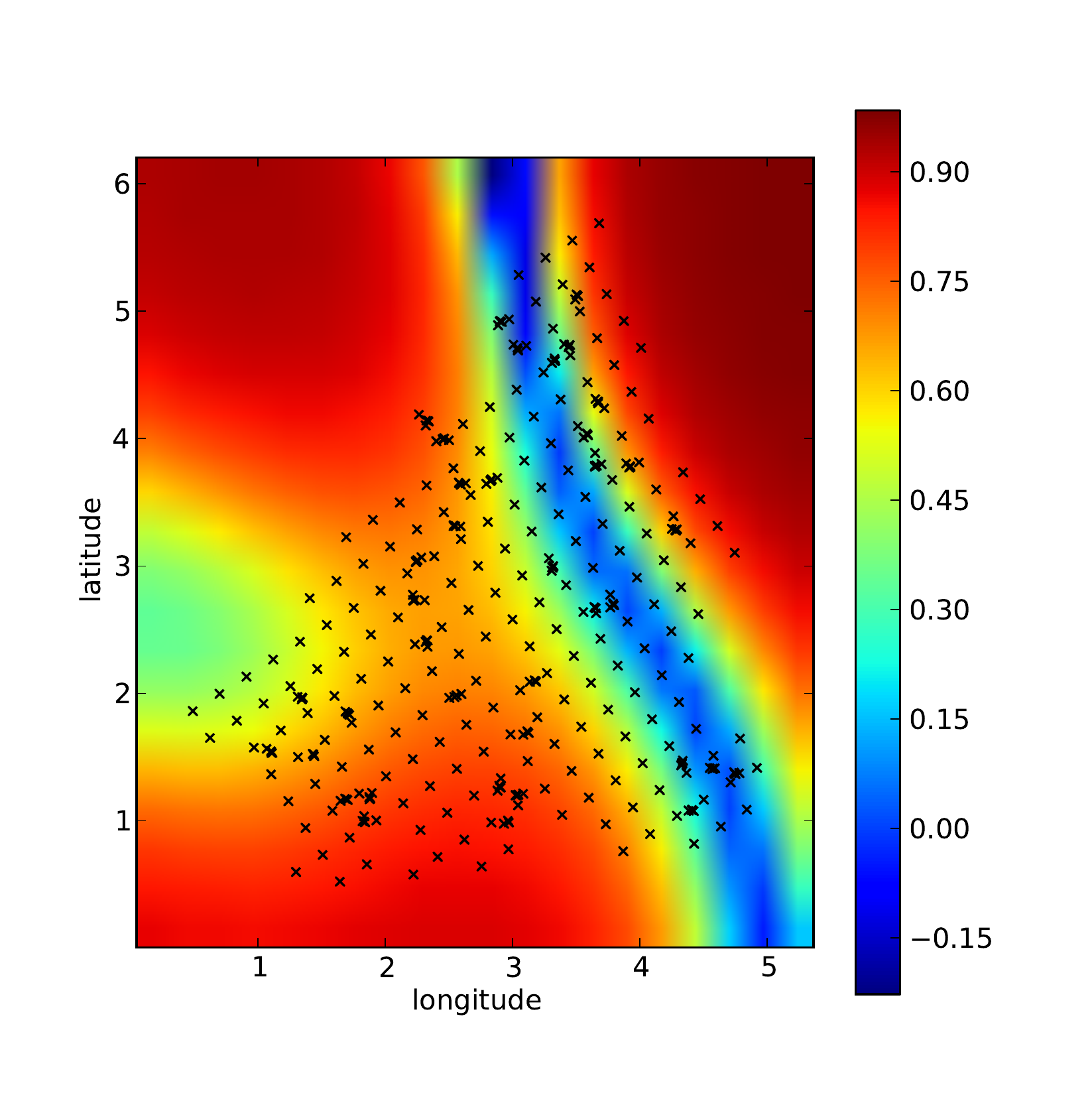}
\caption[Hallo]
{\small Spatially dependent correlation between cadmium and zinc learnt by the GPRN. Markers show the locations where measurements were made. }
\label{fig:juracorrelations}
\end{figure}

Here the network structure with by far the highest marginal likelihood has $q=2$ latent 
node functions. The node and weight functions learnt using VB for this setting are shown in Figure~\ref{fig:juranetwork}. Since there are $p=3$ output dimensions, the result $q<p$ 
suggests that heavy metal concentrations in the Swiss Jura are correlated. Indeed, using our model we can observe the \emph{spatially varying} correlations between heavy metal concentrations, as shown for cadmium and zinc in Figure~\ref{fig:juracorrelations}. 
Although the correlation between cadmium and zinc is generally positive (with values around 0.6), there is a region where the correlations drop of noticeably, perhaps
corresponding to a geological structure.  The quantitative results in Table 1 confirm that the ability of GPRN to learn 
these spatially varying correlations is beneficial in terms of being able to predict cadmium concentrations. 

We assess performance quantitatively using mean absolute error (MAE) between the predicted and true 
cadmium concentrations.  We restart the experiment 10 times with different initialisations 
of the parameters, and average the MAE.  The results are marked by \texttt{JURA} in 
Table \ref{tab: predictions}.  This experiment follows \citet{goovaerts1997} and 
\citet{alvarez2011}.  The results for SLFM, ICM and CMOGP are from \citet{alvarez2011},
and the results for co-kriging are from \citet{goovaerts1997}. It is unclear what preprocessing was performed for these methods, but we found log transforming and normalising each dimension to have zero mean and unit variance to be beneficial due to the skewed distribution of the $y$-values (but we also include results on untransformed data, marked with *). All of the multiple output methods give lower MAE than using an independent GP, and GPRN outperforms SLFM and the other multiple output methods.  

For the \texttt{JURA} dataset, the improved performance of GPRN is at the cost of a slightly greater runtime.
However, GPRN is accounting for input dependent signal and noise correlations, unlike the other methods.
Moreover, the complexity of GPRN scales with $p$ as $\mathcal{O}(Nqp)$, unlike the other methods which scale as 
$\mathcal{O}(N^3 p^3)$ \citep{alvarez2011}.  This is why GPRN runs relatively quickly on the 1000 dimensional gene expression
dataset, for which the other methods are intractable.  This data is available from \url{http://www.ai-geostats.org/}.

\subsection{Multivariate Volatility}
In the previous experiments the GPRN implicitly accounted for multivariate volatility (input dependent noise covariance) in making
predictions of $\bm{y}(x_*)$.  The GPRN incorporates a generalised Wishart process \citep{wilson2010gwp, wilson2011gwp} noise model into a more general model
which can also account for signal correlations and other nonstationarities. 
Although our focus is the ability of GPRN to model input dependent correlations in a multiple output regression setting, 
we here test the GPRN explicitly as a model of multivariate volatility, and assess 
predictions of $\Sigma(t) = \text{cov}[\bm{y}(t)]$, where the observations $\bm{y}$ are time dependent.  We make 
historical predictions at observed time points, and one day ahead forecasts.  Historical
predictions can be used, for example, to understand a past financial crisis.  We follow \citet{wilson2010gwp} exactly, and 
predict $\Sigma(t)$ for returns on three currency exchanges (\texttt{EXCHANGE}) and five equity indices (\texttt{EQUITY}) 
processed exactly as in \citet{wilson2010gwp}.  These datasets are especially suited to MGARCH, the most popular multivariate 
volatility model, and have become a benchmark for assessing GARCH models \citep{granger05, hansen05, engle2009, mccullough98, brooks2001}.  
We make 200 historical predictions of $\Sigma(x)$ and 200 one step ahead forecasts.  The forecasts are assessed using the log likelihood of the new observations under the predicted covariance, denoted $\mathcal{L}$ Forecast. We compare to full BEKK MGARCH \citep{engle1995}, the generalised 
Wishart process \citep{wilson2010gwp}, the original Wishart process \citep{bru91,gourieroux09}, and using the empirical covariance of
the training set.  We see in Table \ref{tab: predictions} that GPRN (VB) is competitive with MGARCH, even though these datasets are 
particularly suited to MGARCH.  The Historical MSE for \texttt{EXCHANGE} is between the learnt covariance $\Sigma(x)$ and $\mathbf{y}(x)\mathbf{y}(x)^{\top}$, so the high MSE values for GPRN on \texttt{EXCHANGE} are essentially training error, and are less meaningful than the encouraging step ahead forecast likelihoods.  The historical
predictions are more relevant in \texttt{EQUITY}, where we can compare to the true covariances.  See \citet{wilson2010gwp} for details. 

GPRN and GWP are both highly flexible but fully Bayesian models for multivariate volatility, so it understandable that their performance is comparable. 
While GPRN (MCMC) sometimes outperforms MGARCH, the GWP, and the WP, it is often outperformed by GPRN (VB) on the multivariate
volatility, perhaps suggesting convergence problems. 
These data were obtained using Bloomberg (\url{http://www.bloomberg.com/}).

\begin{table}
\caption{Comparative performance on all datasets.}
\small
\begin{center}
\begin{tabular}{l r r r r r}
\toprule
\texttt{GENE} (50D)  & Average SMSE  & Average MSLL \\
\midrule             
 \texttt{SET 1}: &   &   \\  
  GPRN (VB) & $0.3356 \pm 0.0294$ & $\bm{-0.5945 \pm 0.0536}$ \\
  GPRN (MCMC) &\bm{ $0.3236 \pm 0.0311 }$ & $-0.5523 \pm 0.0478$  \\
  LMC & $0.6909 \pm 0.0294$ & $-0.2687 \pm 0.0594$ \\
  CMOGP & $0.4859 \pm 0.0387$ & $-0.3617 \pm 0.0511$ \\
  SLFM & $0.6435 \pm 0.0657 $ & $-0.2376 \pm 0.0456$ \\
\midrule
  \texttt{SET 2}: &  &    \\   
  GPRN (VB) & $0.3403 \pm 0.0339$ & $\bm{-0.6142 \pm 0.0557}$ \\
  GPRN (MCMC) & $\bm{0.3266 \pm 0.0321}$ & $-0.5683 \pm 0.0542$  \\
  LMC & $0.6194 \pm 0.0447$ & $-0.2360 \pm 0.0696$ \\
  CMOGP & $0.4615 \pm 0.0626$ & $-0.3811 \pm 0.0748$ \\
  SLFM & $0.6264 \pm 0.0610$ & $-0.2528 \pm 0.0453$ \\
\toprule
\texttt{GENE} (1000D)  & Average SMSE  & Average MSLL \\
\midrule             
  GPRN (VB) & $\bm{0.3473 \pm 0.0062}$ & $\bm{-0.6209 \pm 0.0085}$ \\
  GPRN (MCMC) & $0.4520 \pm 0.0079$ & $-0.4712 \pm 0.0327$  \\
  MFITC  & $0.5469 \pm 0.0125$ & $-0.3124 \pm 0.0200$ \\
  MPITC  & $0.5537 \pm 0.0136$ & $-0.3162 \pm 0.0206$  \\
  MDTC  & $0.5421 \pm 0.0085$ & $-0.2493 \pm 0.0183$ \\
\toprule
  \texttt{JURA} &  Average MAE & Training Time (secs) \\
\midrule 
  GPRN (VB) & $\bm{0.4040  \pm  0.0006}$ & $3781$ \\
  GPRN* (VB) & $ 0.4525 \pm 0.0036$ & $4560$ \\
  SLFM (VB) & $0.4247  \pm  0.0004$ & $1643$ \\
  SLFM* (VB) & $ 0.4679  \pm  0.0030$ & $1850$ \\
  SLFM & $0.4578 \pm 0.0025$  & $792$ \\ 
  Co-kriging &  0.51 & \\
  ICM & $0.4608 \pm 0.0025$ & $507$ \\
  CMOGP & $0.4552 \pm 0.0013$ & $784$ \\
  GP & $0.5739 \pm 0.0003$ & $74$ \\
\toprule
  \texttt{EXCHANGE} & Historical MSE & $\mathcal{L}$ Forecast \\
\midrule
  GPRN (VB) & $3.83 \times 10^{-8}$ & $\bm{2073}$ \\ 
  GPRN (MCMC) & $6.120 \times 10^{-9}$ & $2012$ \\
  GWP & $\bm{3.88 \times 10^{-9}}$ & $2020$ \\
  WP & $\bm{3.88 \times 10^{-9}}$ & $1950$ \\
  MGARCH & $3.96 \times 10^{-9}$ & $2050$ \\
  Empirical & $4.14 \times 10^{-9}$ & $2006$ \\
\toprule
  \texttt{EQUITY} & Historical MSE & $\mathcal{L}$ Forecast  \\
\midrule
  GPRN (VB) & $0.978 \times 10^{-9} $ & $2740$ \\
  GPRN (MCMC) & $\bm{0.827 \times 10^{-9}}$ & $2630$   \\
  GWP & $2.80 \times 10^{-9}$ & $\bm{2930}$\\
  WP  & $3.96 \times 10^{-9}$ & $1710$\\
  MGARCH & $6.69 \times 10^{-9}$ & $2760$ \\
  Empirical & $7.57 \times 10^{-9}$ & $2370$ \\
\bottomrule
\end{tabular}
\end{center}
\label{tab: predictions}
\end{table}

\section{Discussion}
A Gaussian process regression network (GPRN) has a simple and interpretable 
structure, and generalises many of the recent extensions to the Gaussian 
process regression framework.  The model naturally accommodates input dependent 
signal and noise correlations between multiple output variables, heavy tailed    
predictive distributions, input dependent length-scales and amplitudes, and 
adaptive covariance functions.  Furthermore, GPRN has scalable inference
procedures, and strong empirical performance on several benchmark datasets.

In the future, it would be enlightening to use GPRN with different types of 
adaptive covariance structures, particularly in the case where $p=1$ and $q>1$; 
in one dimensional output space it would be easy, for instance, to visualise a process gradually 
switching between brownian motion, periodic, and smooth covariance functions.  It would also be 
interesting to apply this adaptive network to classification.  We hope the GPRN will 
inspire further research into adaptive networks, and further connections between different areas of 
machine learning and statistics.

\section{Acknowledgements}
Thanks to Mauricio \'{A}lvarez and Neil Lawrence for their valuable feedback about the 
gene expression datasets.

\section{Appendix}

\subsection{Gaussian processes}  
Since we extend the Gaussian process framework, we briefly review Gaussian process regression, some notation,
and expand on some of the points in the introduction.  For more detail see \citet{rasmussen06}.

A Gaussian process is a collection of random variables, any finite number of which have a joint Gaussian 
distribution.  Using a Gaussian process, we can define a distribution over functions $w(x)$:
\begin{equation}
 w(x) \sim \mathcal{GP}(m(x),k(x,x')),
\end{equation}
where $w$ is the output variable, $x$ is an arbitrary (potentially vector valued) input variable, and
the mean $m(x)$ and covariance function (or kernel) $k(x,x')$ are respectively defined as
\begin{align}
m(x) &= \mathbb{E}[w(x)] \,, \\
k(x,x') &=  \text{cov}[w(x),w(x')]\,.
\end{align}
This means that any collection of function values has a joint Gaussian distribution:
\begin{equation}
(w(x_1),w(x_2),\dots,w(x_N))^{\top} \sim \mathcal{N}(\bm{\mu},K) \,, \label{eqn: joint}
\end{equation}
where the $N \times N$ covariance matrix $K$ has entries $K_{ij} = k(x_i,x_j)$, and
the mean $\bm{\mu}$ has entries $\bm{\mu}_i = m(x_i)$.  The properties of these
functions (smoothness, periodicity, etc.) are determined by the kernel function.  The
squared exponential kernel is popular:
\begin{equation}
 k_{\text{SE}}(x,x') = A \exp(-0.5||x-x'||^2/l^2) \,. \label{eqn: squaredexp}
\end{equation}
Functions drawn from a Gaussian process with this kernel function are smooth, and
can display long range trends.  The length-scale \textit{hyperparameter} $l$ is easy
to interpret: it determines how much the function values $w(x)$ and
$w(x+a)$ depend on one another, for some constant $a \in \mathcal{X}$.  When the
length-scale is learned from data, it is useful for determining how far into 
the past one should look in order to make good forecasts.  $A \in \mathbb{R}$
is the \textit{amplitude} coefficient, which determines the marginal variance
of $w(x)$ in the prior, $\text{Var}[w(x)] = A$, and the magnitude of covariances
between $w(x)$ at different inputs $x$.

The Ornstein-Uhlenbeck kernel is also widely applied:
\begin{equation}
k_{\text{OU}}(x,x') = \exp(-||x-x'||/l) \,.
\end{equation}
In one dimension it is the covariance function of an Ornstein-Uhlenbeck process \citep{uhlenbeck30}, 
which was introduced to model the velocity of a particle undergoing Brownian motion. With this kernel, the 
corresponding GP is a continuous time AR(1) process with Markovian dynamics: $w(x+a)$ is independent of 
$w(x-a)$ given $w(x)$ for any constant $a$.  Indeed the OU kernel belongs to a more general class of 
Mat\'{e}rn kernels,
\begin{equation} 
k_{\text{Mat\'{e}rn}}(x,x') = \frac{2^{1-\alpha}}{\Gamma(\alpha)}(\frac{\sqrt{2\alpha}||x-x'||}{l})^{\alpha}K_{\alpha}(\frac{\sqrt{2\alpha}||x-x'||}{l})\,, \label{eqn: matern}
\end{equation}
where $K_{\alpha}$ is a modified Bessel function \citep{abramowitz64}.  In one dimension the corresponding GP 
is a continuous time AR($p$) process, where $p = \alpha + 1/2$.\footnote{Discrete time autoregressive processes such as
$w(t+1) = w(t) + \epsilon(t)$, where $\epsilon(t) \sim \mathcal{N}(0,1)$, are widely used in time series 
modelling and are a particularly simple special case of Gaussian processes.} 
The OU kernel is recovered by setting $\alpha = 1/2$.

There are many other useful kernels, like the periodic kernel (with a period that can be learned from data), or the 
Gibbs kernel \citep{gibbs97} which allows for input dependent length-scales.  Kernels can be combined together, e.g. 
$k = a_1 k_1 + a_2 k_2 + a_3 k_3$, and the relative importance of each kernel can be determined from data 
(e.g. from estimating $a_1, a_2, a_3$).  \citet{rasmussen06} and \citet{bishop06} have a discussion about how to 
create and combine kernels.  

Suppose we are doing a regression using points $\{y(x_1),\dots,y(x_N)\}$ from a noisy function $y = w(x) + \epsilon$, 
where $\epsilon$ is additive i.i.d Gaussian noise, such that $\epsilon \sim \mathcal{N}(0,\sigma_n^2)$.  Letting
$\bm{y} = (y(x_1),\dots,y(x_N))^{\top}$, and $\bm{w} = (w(x_1),\dots,w(x_N)^{\top}$, we have 
$p(\bm{y} | \bm{w}) = \mathcal{N}(\bm{w},\sigma_n^2 I)$ and $p(\bm{w}) = \mathcal{N}(\bm{\mu},K)$ as above.  
For notational simplicity, we assume $\bm{\mu} = 0$.  For a test point $w(x_*)$, the joint distribution 
$p(w(x_*),\bm{y})$ is Gaussian:
\begin{align}
\begin{bmatrix} w(x_*) \\ \bm{w} \end{bmatrix} 
\sim \mathcal{N}(\bm{0},
\left[
\begin {array}{cc}
k(x_*,x_*) & \bm{k}_*^{\top}\\
\noalign{\medskip}
\bm{k}_* & K + \sigma_n^2 I\
\end {array}
\right])\,,
\end{align}
where $K$ is defined as above, and $(\bm{k}_*)_i = k(x_*,x_i)$ with $i=1,\dots,N$.  We can therefore 
condition on $\bm{y}$ to find $p(w(x_*)|\bm{y}) = \mathcal{N}(\mu_*,v_*)$ where
\begin{align}
\mu_* = \bm{k}_*^{\top}(K+\sigma_n^2 I)^{-1} \bm{y} \,, \label{eqn: mustarnew}  \\  
v_* = k(x_*,x_*) - \bm{k}_*^{\top}(K+\sigma_n^2 I)^{-1} \bm{k}_* \,.  \label{eqn: vstar}
\end{align}
We can find this more laboriously by noting that $p(\bm{w}|\bm{y})$ and $p(w(x_*)|\bm{w})$ are Gaussian and
integrating, since $p(w(x_*)|\bm{y}) = \int p(w(x_*) | \bm{w})p(\bm{w}|\bm{y}) d\bm{w}$.

We see that \eqref{eqn: vstar} doesn't depend on the data $\bm{y}$, just on how far away the test point
$x_*$ is from the training inputs $\{x_1,\dots,x_N\}$.  

In regards to the introduction, we also see that for this standard Gaussian process regression, the 
observation model $p(y|w)$ is Gaussian, the predictive distribution in \eqref{eqn: mustarnew}
and \eqref{eqn: vstar} is Gaussian, the marginals in the prior (from marginalising equation
\eqref{eqn: joint}) are Gaussian, the noise is constant, and in the popular 
covariance functions given, the amplitude and length-scale are constant.
A brief discussion of multiple outputs, noise models with dependencies, and 
non-Gaussian observation models can be found in sections 9.1, 9.2 and 9.3 on pages 190-191 
of \citet{rasmussen06}, available free online at the book website \url{www.gaussianprocess.org/gpml}.
An example of an input dependent length-scale is in section 4.2 on page 43.

\subsection{Constraining $W$}

It is possible to reduce the number of modes in the posterior by somehow constraining the weights $W$ to be positive. 
For MCMC it is straightforward to do this by exponentiating the weights, as in~\cite{adams2008} and \cite{adams2010side}. 
For VB it is more straightforward to explicitly constrain the weights to be positive using a truncated Gaussian representation. 
We found that these extensions did not significantly improve empirical performance, although exponentiating the weights sometimes
improved numerical stability for MCMC on the multivariate volatility experiments. For \cite{adams2008} exponentiating the weights 
will have been more valuable because they use Expectation Propagation which is known to perform badly in the presence of multimodality. 
MCMC and VB approaches are more robust to this problem. 

\subsection{VB M-step}

From $\eqref{eqn:gpfactor}$ we have:
\begin{align*}
& \langle \log \mathcal{N}(\mathbf{f}_j;0,a_jK_{f_j}) \rangle_{q} \\
& \stackrel{c}{=} - \frac{N}2 \log a_j -\frac12 \log |K_{f_j} | - \frac12 \langle a^{-1}_j \rangle \langle \mathbf{f}^T_j K^{-1}_{f_j} \mathbf{f}_j \rangle
\end{align*}
We will need the gradient with respect to $\theta_f$:
\begin{align*}
& \frac{\partial \langle \log \mathcal{N}(\mathbf{f}_j;0,a_jK_{f_j}) \rangle}{\partial \theta_f} \\
& = -\frac12 \text{tr} \left(K^{-1}_{f_j} \frac{\partial K_{f_j}}{\partial \theta_f} \right) - \frac12 \langle a^{-1}_j \rangle \langle \mathbf{f}^T_j K^{-1}_{f_j} \frac{\partial K_{f_j}}{\partial \theta_f} K^{-1}_{f_j} \mathbf{f}_j \rangle
\end{align*}
The expectations here are straightforward to compute analytically. 

\subsection{VB predictive distributions}

The predictive distribution is calculated as
\begin{align}
p(\mathbf{y}^*(x)|\mathcal{D}) &= \int p(\mathbf{y}^*(x)|W(x),f(x))p(W(x),f(x)|\mathcal{D}) dWdf
\end{align}
VB fits the approximation $p(W(x),f(x)|\mathcal{D})=q(W)q(f)$, so the approximate predictive is
\begin{align}
p(\mathbf{y}^*(x)|\mathcal{D}) &= \int p(\mathbf{y}^*(x)|W(x),f(x))q(W)q(f) dWdf
\end{align}
We can calculate the mean and covariance of this distribution analytically:
\begin{align}
\bar{\mathbf{y}}^*(x)_i &= \sum_k \mathbb{E}(W_{ik}) \mathbb{E}[f_k] \\
\text{cov}(\mathbf{y}^*(x))_{ij} &= \sum_k [\mathbb{E}(W_{ik})\mathbb{E}(W_{jk})\text{var}(f_k) + \delta_{ij}(\text{var}(W_{ik})\mathbb{E}(f_k^2))] + \delta_{ij}\sigma_y^2
\end{align}
It is also of interest to calculate the \emph{noise} covariance. Recall our model can be written as
\begin{align}
 \bm{y}(x) = \underbrace{W(x)\bm{f}(x)}_{\text{signal}} + \underbrace{\sigma_f W(x)\bm{\epsilon} + \sigma_y \bm{z}}_{\text{noise}}  
\end{align}
Let $\bm{n}=\sigma_f W(x)\bm{\epsilon} + \sigma_y \bm{z}$ be the noise. The covariance of $\mathbf{n}$ is then
\begin{align}
\text{cov}(\bm{n})_{ij} &= \sum_k [\mathbb{E}[\sigma^2_{f_k}] \mathbb{E}(W_{ik})\mathbb{E}(W_{jk}) + \delta_{ij} \text{var}(W_{jk})] + \delta_{ij} \sigma_y^2
\end{align}

\bibliographystyle{apalike} 
\bibliography{mbib}

\end{document}